\documentclass[runningheads]{llncs}
\usepackage{graphicx}
\usepackage{amsmath}  %
\usepackage{amssymb}  %

\usepackage{tikz}
\usetikzlibrary{shapes.misc, positioning, arrows.meta}
\tikzset{
layer/.style={
    rounded rectangle,
    fill=gray!25!white,
    draw=white,
    minimum width=1.5cm
}
}
\newcommand{\appendixcontext}[2]{#2}  %

\usepackage{siunitx}
\usepackage{multirow}
\usepackage{booktabs}

\title{Siloed Federated Learning for Multi-Centric
Histopathology Datasets}
\author{Mathieu Andreux\inst{1} \and
Jean Ogier du Terrail\inst{1} \and\\
Constance Beguier \and
Eric W. Tramel}
\authorrunning{Andreux et al.}
\institute{Owkin, Inc., New York, NY, USA \\
\email{firstname.lastname@owkin.com}}
\begin{document}
\maketitle              %
\footnotetext[1]{Contributed equally to this work.}
\begin{abstract}
While federated learning is a promising approach for training
deep learning models over distributed sensitive
datasets, it presents new challenges for machine learning, 
especially when applied in the medical domain where 
multi-centric data heterogeneity is common.
Building on previous domain adaptation works,
this paper proposes 
a novel federated learning approach
for deep learning architectures
via the introduction of local-statistic batch normalization (BN) layers,
resulting in collaboratively-trained, yet center-specific models.
This strategy improves robustness to data heterogeneity
while also reducing the potential for information leaks by not sharing
the center-specific layer activation statistics.
We benchmark the proposed method on the classification
of tumorous histopathology image patches extracted from
the Camelyon16 and Camelyon17 datasets.
We show that our approach
compares favorably to previous state-of-the-art methods,
especially for transfer learning across datasets.
\end{abstract}
\section{Introduction}\label{sec:intro}
Federated learning (FL) has recently emerged
as a new paradigm for scalable and practical
privacy-preserving machine learning (ML)
on decentralized datasets~\cite{MMR2017}.
In the case of medical data, notably digital histopathology images,
this approach brings the promise of ML architectures trained over
large and diverse populations, a necessary component for truly generalizable
medical findings. By bridging the gap between localized,
curated, and non-portable per-institution datasets, a 
federated approach permits 
the study of otherwise unconstructible
research datasets. 
With the ability to utilize large and rich datasets for ML,
medical researchers have the potential to make new scientific 
discoveries, as evidenced in~\cite{CMM2019}.

Two challenges currently limit the applicability of existing 
FL techniques to real-world histopathology datasets:
inter-center data heterogeneity and well-understood privacy assurances 
on communicated model parameters.
Statistical heterogeneity in the data distribution
between participating centers is a key issue for FL~\cite{SLS2018,KMA2019,LST2019},
which may lead to model biases or even prevent training convergence~\cite{SLS2018}.
Such heterogeneity is often present in histopathology datasets,
where variations in staining procedure, scanning device configuration,
and systematic imaging artifacts are commonplace~\cite{KI2018}.

In this paper, we propose 
a novel federated learning strategy,
called \textit{SiloBN},
which brings improvements both in terms
of resilience to heterogeneity and privacy.
This strategy relies on batch normalization (BN)~\cite{IS2015} layers,
which are ubiquitous in deep learning (DL), especially for computer vision applications.
\textit{SiloBN} introduces, or uses already specified, 
\emph{local-statistic} BN layers within
DL architectures.
More precisely, 
we propose the  following novel contributions:
\begin{itemize}
    \item[\emph{i})]  We demonstrate the applicability of FL
to real-world tile-level digital pathology image classification,
using the Camelyon16 and Camelyon17
datasets~\cite{LBB2018} (Sec.~\ref{sec:experiments}).
To the best of our knowledge, this work is among the first examples thereof.
Previous works rather relied on the SplitLearning
framework~\cite{vepakomma2019splitlearning}.
\item[\emph{ii})] We introduce \textit{SiloBN}, a new FL approach
for training DL models robust to inter-center data
variability by introducing local-statistic BN layers. 
This approach also reveals less sensitive information
by not communicating local activation statistics (Sec.~\ref{sec:siloedbn}).
\item[\emph{iii})]%
We show that our proposed approach achieves same or better performance
than existing federated techniques for intra-center generalization
(Sec.~\ref{sec:silosxps}), even
in challenging settings,
and yields better out-of-domain generalization results
(Sec.~\ref{sec:domaingeneralizationxps}).
\end{itemize}

\section{Background}%
\label{sec:background}

In the years since the publication of the 2013 ICPR-winning
approach of~\cite{CGG2013}, the application of DL to digital pathology
tasks has blossomed~\cite{DAC2019,vepakomma2019splitlearning}.
In parallel, federated learning~\cite{SS2015,MMR2017},
a distributed privacy-preserving ML paradigm,
has recently experienced incredible growth as a topic of study,
garnering much interest in the medical research community.
Due to the highly sensitive nature of the medical data, techniques such as FL
are a requirement for investigators in order to develop
state-of-the-art ML models over a fractured and highly regulated
data landscape.
\paragraph{Federated Learning}
As presented in~\cite{MMR2017}, 
the aim of FL is to obtain a single well-trained model
from a distributed network of $K$ participants, each
possessing their own privately-held datasets.
Generically,
this is accomplished by optimizing model parameters
$\theta$ w.r.t. an expectation over
the individual participant losses~$\ell_i$,
i.e.~$\mathcal{L}(\theta)\triangleq \mathbb{E}_i\left[ \ell_i(\theta)\right]$.
This optimization is carried out iteratively in federated rounds,
where at each round $t$ the expectation is approximated
as~$\sum_{i\in \mathcal{C}^{(t)}} \alpha_i\ell_i(\theta)$,
where~$\mathcal{C}^{(t)} \sim [K]$ is random sub-sampling of participants
and~$\alpha_i \geq 0$ are contribution weightings.
The authors propose \emph{federated averaging} (\textit{FedAvg})
to reduce coordination rounds by performing several local
optimization steps prior to aggregation,
\begin{align}
    \overbrace{\theta_{i}^{(t)} = \text{Opt}_{E}\left( \theta^{(t)}, \ell_i\right)}^{\text{Local}},
    \quad\quad\quad
    \overbrace{%
        \theta^{(t+1)} = 
        \sum_{i\in\mathcal{C}^{(t)}} \frac{N_i}{N^{(t)}} \theta_i^{(t)}}^{\text{Central Server},
    },
    \label{eq:fedavg}
\end{align}
where $N_i$ is the number of data samples at participant $i$,
and $N^{(t)} \triangleq \sum_{i\in\mathcal{C}^{(t)}} N_i$,
$\text{Opt}_E(\cdot)$ is
an $E$-step iterative optimization, and $t$ indexes federated rounds.
For brevity, we refer the reader
to the original work of~\cite{MMR2017} for details. 
Further adaptations can be
made to the algorithm to enhance privacy, 
e.g.~\cite{ACG2016,BIK2017}, but are out of scope of the present work.

\paragraph{Data Heterogeneity in FL}
Data heterogeneity has been identified as a key open challenge
for FL~\cite{LST2019}.
For example, despite its practical success, \textit{FedAvg} does not provide a 
natural guarantee of convergence, especially for highly 
dissimilar participant datasets. The \textit{FedProx} algorithm of
~\cite{SLS2018}, similar to the \textit{EASGD} of \cite{ZCL2015},
introduces a quadratic loss term at each round~$\ell_i^{(t)}(\theta_i)=\ell_i(\theta_i) + \frac{\lambda}{2}||\theta_i - \theta^{(t)}||^2$, 
to restrain local models from diverging during local optimization.
This approach was shown to be effective on the large-$K$, heterogeneous
LEAF datasets~\cite{CWL2018}. 
Novel participant sampling strategies have also 
been recently proposed~\cite{LHY2019,GMB2019} to
help adapt \textit{FedAvg} to 
heterogeneous participant datasets. Finally, one
can also seek to partition the set of participants into clusters
of effective collaborators, as in~\cite{SMS2019}.
It should be noted that the assumption that data samples
are independent and identically distributed (\emph{i.i.d.})
is core to many ML
techniques, and empirical risk minimization~\cite{V1999} in particular.
For FL, we retain an \emph{i.i.d.} assumption on data
\emph{intra-}participant, but not \emph{inter-}participant.

\section{FL in Healthcare}
The original application of FL sought to define techniques
for distributed training over mobile edge devices, e.g. smart phones
~\cite{KMR2016,MMR2017}, and has now been practically scaled to 
$K > 10^6$~\cite{BEG2019}. Notably, this application also assumes that each
participant has a paucity of training data,
i.e.~$N_i \ll N = \sum_{j=1}^{K} N_j~~\forall i\in [K]$.
The application of FL to the context of distributed and private
\emph{medical datasets} reverses the order of magnitude on both of these
variables, with the number of participants often being quite low,
e.g. $K \leq 10$, and generally having access to much larger and higher 
quality datasets at each center, i.e. $\log N_i \approx \log N$.
The implication of this reversal is 
that the nature and goals of the federated system shift between the
two settings:
\begin{enumerate}
    \item[i)] \emph{Consumer FL} --- Federated training coordinated and of value 
        to central service provider. No single participant, out of
        very many, can train
        an effective model, and no participant has an incentive to use
        such a local model. Goal of the service provider is to coordinate
        and distribute final global model back to participants. Service 
        provider may hold their own private or public evaluation data.

    \item[ii)] \emph{Collaborative FL} --- Federated training coordinated by
        central service provider \emph{on behalf} of few participants, who
        value and seek to reuse its outputs. Participants have enough
        local data of quality to produce effective local models, but
        seek to augment this local task performance through 
        collaboration. Evaluation data is sensitive and held by participants.
\end{enumerate}
For this work, we consider the \emph{collaborative FL setting}
for training  a tile-level classifier for WSI across multiple centers. 
This setting poses several challenges for the construction of
distributed algorithms which are not fully addressed in the literature
described in Sec.~\ref{sec:background}.

Data heterogeneity is a key challenge for collaborative FL, as
inconsistencies in data preparation will not be averaged out over a 
large population of participants. Additionally, in the consumer FL 
setting, $|\mathcal{C}(t) | \ll K$, which means outlier participants are
rarely revisited. For collaborative FL, such strong sub-sampling is not
desirable, as per-round updates would become \emph{more} biased and also
lead to potential privacy challenges as the sensitivity of the aggregated
model w.r.t. the center of origin increases. For this reason, we consider
only full-participation federated rounds in the case of collaborative FL.
In order to keep stochasticity in the optimization, contrary
to~\cite{MMR2017} which uses local epochs, we use $E$ to denote
the number of local \textit{mini-batch} updates in~\eqref{eq:fedavg}.
\section{Proposed Method}%
\label{sec:siloedbn}

\paragraph{Batch Normalization (BN)}
BN~\cite{IS2015} is a critical component
of modern deep learning architectures.
A BN layer renormalizes an input tensor $x$ as
\begin{equation}
    \text{BN}(x) =  
    \gamma \frac{x - \mu}{\sqrt{\sigma^2 + \epsilon}} + \beta,
    \label{eq:bn}
\end{equation}
where $\mu$ and $\sigma^2$,
hereafter denoted as BN statistics,
are calculated as the running means and variances, respectively,
of each channel computed across both spatial and batch dimensions,
and $\gamma$ and $\beta$ are \emph{learned} affine renormalization
parameters, and where all computations
are performed along the channel axis.
BN has been used to both speed model training and also enhance model
predictive performance \cite{IS2015}. 

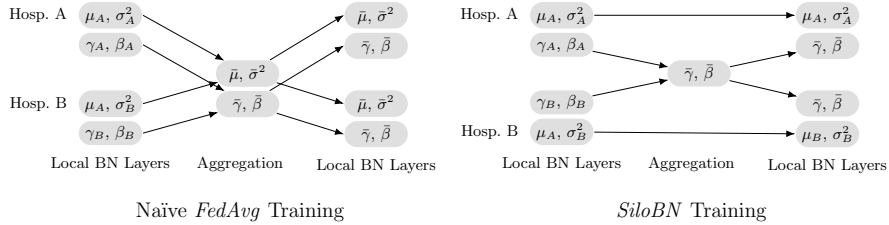
\begin{figure}[ht!]
\resizebox{!}{3cm}{%
\begin{tikzpicture}
    \node (hospA_t_stats) [layer] {$\mu_A$, $\sigma^2_A$};
    \node (hospA_t_params) [layer, below = 0.05cm of hospA_t_stats] {$\gamma_A$, $\beta_A$};
    \node (hospB_t_stats) [layer, below = 1.2cm of hospA_t_stats] {$\mu_A$, $\sigma^2_B$};
    \node (hospB_t_params) [layer, below = 00.05cm of hospB_t_stats] {$\gamma_B$, $\beta_B$};

    \node (agg_stats) [layer, below right = 0.6cm and 2cm of hospA_t_stats] {$\bar{\mu}$, $\bar{\sigma}^2$};
    \node (agg_params) [layer, below = 0.05cm of agg_stats] {$\bar{\gamma}$, $\bar{\beta}$};

    \node (hospA_tp_stats) [layer, right = 4cm of hospA_t_stats] {$\bar{\mu}$, $\bar{\sigma}^2$};
    \node (hospA_tp_params) [layer, below = 0.05cm of hospA_tp_stats] {$\bar{\gamma}$, $\bar{\beta}$};
    \node (hospB_tp_stats) [layer, below = 1.2cm of hospA_tp_stats] {$\bar{\mu}$, $\bar{\sigma}^2$};
    \node (hospB_tp_params) [layer, below = 0.05cm of hospB_tp_stats] {$\bar{\gamma}$, $\bar{\beta}$};

    \node (hospA) [left = 0.1cm of hospA_t_stats] {Hosp. A};
    \node (hospB) [left = 0.1cm of hospB_t_stats] {Hosp. B};

    \node (local_t) [below = 0.1cm of hospB_t_params] {Local BN Layers};
    \node (agg) [right = 0.3cm of local_t] {Aggregation};
    \node (local_tp) [below = 0.1cm of hospB_tp_params] {Local BN Layers};

    \draw[-Latex] (hospA_t_stats.east) -- (agg_stats);
    \draw[-Latex] (agg_stats) -- (hospA_tp_stats.west);

    \draw[-Latex] (hospA_t_params.east) -- (agg_params);
    \draw[-Latex] (agg_params) -- (hospA_tp_params.west);
    \draw[-Latex] (hospB_t_stats.east) -- (agg_stats);
    \draw[-Latex] (agg_stats) -- (hospB_tp_stats.west);

    \draw[-Latex] (hospB_t_params.east) -- (agg_params);
    \draw[-Latex] (agg_params) -- (hospB_tp_params.west);

    \node (title) [below = 0.4cm of agg] {\large{Na\"ive \textit{FedAvg} Training}};

\end{tikzpicture}   
 }
\resizebox{!}{3cm}{%
\begin{tikzpicture}
    \node (hospA_t_stats) [layer] {$\mu_A$, $\sigma^2_A$};
    \node (hospA_t_params) [layer, below = 0.05cm of hospA_t_stats] {$\gamma_A$, $\beta_A$};

    \node (hospB_t_params) [layer, below = 1.2cm of hospA_t_stats] {$\gamma_B$, $\beta_B$};
    \node (hospB_t_stats) [layer, below = 0.05cm of hospB_t_params] {$\mu_A$, $\sigma^2_B$};

    \node (agg_params) [layer, below right = 0.6cm and 2cm of hospA_t_stats] {$\bar{\gamma}$, $\bar{\beta}$};

    \node (hospA_tp_stats) [layer, right = 4cm of hospA_t_stats] {$\mu_A$, $\sigma^2_A$};
    \node (hospA_tp_params) [layer, below = 0.05cm of hospA_tp_stats] {$\bar{\gamma}$, $\bar{\beta}$};

    \node (hospB_tp_params) [layer, below = 1.2cm of hospA_tp_stats] {$\bar{\gamma}$, $\bar{\beta}$};
    \node (hospB_tp_stats) [layer, below = 0.05cm of hospB_tp_params] {$\mu_B$, $\sigma^2_B$};

    \node (hospA) [left = 0.1cm of hospA_t_stats] {Hosp. A};
    \node (hospB) [left = 0.1cm of hospB_t_stats] {Hosp. B};

    \node (local_t) [below = 0.1cm of hospB_t_stats] {Local BN Layers};
    \node (agg) [right = 0.3cm of local_t] {Aggregation};
    \node (local_tp) [below = 0.1cm of hospB_tp_stats] {Local BN Layers};

    \draw[-Latex] (hospA_t_stats) edge  (hospA_tp_stats);
    \draw[-Latex] (hospA_t_params) edge (agg_params) (agg_params) edge (hospA_tp_params);
    \draw[-Latex] (hospB_t_stats) edge  (hospB_tp_stats);
    \draw[-Latex] (hospB_t_params) edge (agg_params) (agg_params) edge (hospB_tp_params);

    \node (title) [below = 0.4cm of agg] {\large{\emph{SiloBN} Training}};

\end{tikzpicture}   
 }
    \caption{\label{fig:silobndesc}%
        Description of the different approaches to multi-center training of 
        BN layers for two hospitals. In this description, all variables follow
        the definitions given in~\eqref{eq:bn}.
        Computation flows from left to right.
        Non-BN layers are shared in both methods.
    }
\end{figure}

While BN layers are common architectural
components~\cite{he2016residual,hu2018squeeze,tan2019efficient},
they have not been thoroughly addressed 
in the federated setting, and are often simply ignored or removed~\cite{MMR2017}.
Indeed, the na\"ive application of \textit{FedAvg} would ignore the different
roles of the local activation statistics $(\mu, \sigma^2)$, and the
trained renormalization $(\gamma, \beta)$ and simply aggregate both
at each federated round, as depicted in Fig.~\ref{fig:silobndesc} (left).
In the following, we use this method as a \textit{baseline}
when using \textit{FedAvg} on a network with BN layers.

BN layers can also be used to separate local and domain-invariant
information.
Specifically, the BN statistics and learned BN parameters 
play different roles~\cite{LWS2016}:
while the former encode \emph{local} domain information,
the latter can be transferred across domains. This opens interesting
possibilities to tune models to local datasets, which we discuss
in the following section.

\paragraph{SiloBN}
Instead of treating all BN parameters equally, we propose
to take into account the separate roles of BN statistics~$(\mu, \sigma^2)$ and
learned parameters~$(\gamma, \beta)$.
Our method, called \textit{SiloBN}, consists in only
sharing the learned BN parameters across different centers,
while \emph{BN statistics remain local}.
Parameters of non-BN layers are shared in the standard fashion.
This method is depicted in Fig.~\ref{fig:silobndesc} (right).
Keeping BN statistics local permits the federated training
of a model robust to the heterogeneity of the different centers,
as the \emph{local statistics ensure that the intermediate
activations are centered to a similar value across centers}. While
\emph{SiloBN} can be applied to models which already possess BN layers, we
demonstrate that in some cases, BN layers can be added to the base model
to improve resilience to heterogeneity as well as domain generalization.

\paragraph{Model Personalization and Transfer}
A consequence of the proposed method is that one model
per center is learned instead of a single global model,
thereby enabling model personalization.
However, it is not straightforward to generalize the resulting models
to unseen centers, as BN statistics must be tuned to the target dataset.
A simple approach to overcome this issue is to follow
the \textit{AdaBN}~\cite{LWS2016} method, and recompute BN statistics on a
data batch of the new target domain, while all other model parameters
remain frozen to those resulting from the federated training.

\paragraph{Privacy and BN}
When applying standard FL methods on a network containing
BN layers, the BN statistics are shared between centers and may
leak sensitive information from local training datasets.
With \textit{SiloBN}, the 
BN statistics are not shared among centers, and in particular
the central server never has access to all the network parameters.
As a number of privacy attacks~\cite{shokri2017membership,ZLH2019}
are most effective with white-box, full-parameter, knowledge,
the reduction of the amount of shared information can only help diminish
the effectiveness of such attacks when compared to
synchronizing batch activation statistics across participants.
\section{Experiments}
\label{sec:experiments}
\subsection{Datasets}
\label{subsec:datasets}
In order to understand the performance of federated algorithms
in a practical setting, we seek real-world datasets with actual
multi-centric origins.
For histopathology, the Camelyon~16~\cite{BVV2017} and
Camelyon~17~\cite{BGM2018} challenge datasets
provide H\&E stained whole slide images (WSI) of lymph node sections 
drawn from breast cancer patients from 2 and 5 hospitals, respectively.
Fig.~\appendixcontext{\ref{fig:cam-slides}}{4} in Supp. Mat.
shows representative slides from each
center, which have different color statistics on average.

Building on Camelyon~16 and Camelyon~17, we construct two tile-level tumor classification
histopathology datasets, referred to as FL-C16 and FL-C17.
In both cases,
the task consists in classifying the tiles between tumorous
and healthy ones.
In order to get tile-level annotations, tumorous tiles
are extracted from pixel-level annotated tumorous WSI, whenever available,
while healthy tiles are extracted from healthy WSI.
This restriction, motivated by incomplete expert annotations, ensures tile-level labels' correctness.
For each WSI, we extract at most~$10,000$ non-overlapping
fixed-size matter-bearing tiles uniformly at 
random using a U-Net segmentation model~\cite{RFB2015} trained for matter detection. 
To reduce class imbalance, for each healthy WSI, we cap the number of 
extracted tumor-negative tiles to~$1,000$ for FL-C16 and~$100$ for FL-C17,
as FL-C17 has fewer annotated tumor-positive tiles than FL-C16.
Finally, each dataset is partitioned into a training set~(60\%), a validation set (20\%), 
and a test set~(20\%) using per-hospital stratification.
For FL-C16 (resp. FL-C17), tiles 
from same slide (resp. tiles from same patient) are put into the same partition. 
Table~\appendixcontext{\ref{table:training_set_distribution}}{3}
in Supp. Mat. details the 
distribution of tiles in the data partitions.

\subsection{Experiments on FL Histopathology Datasets}
\label{sec:silosxps}

\paragraph{Baselines}
We compare the proposed \textit{SiloBN} method to two standard Federated
algorithms,
\textit{FedAvg} and \textit{FedProx},
which treat BN statistics as standard parameters.
Additionally, we report results obtained in the \textit{pooled} setting, where 
the training sets of each center are concatenated and the origin center information
is discarded during training.
To measure the benefits of the personalization of our model, we also compare 
the proposed method to a \textit{local} training,
where on each center one model is trained independently without inter-center collaboration.

\begin{figure}[t!]
\begin{center}
        \includegraphics[width=0.7\textwidth, trim={0.2cm, 0.3cm, 0.2cm, 0.2cm}, clip]{%
        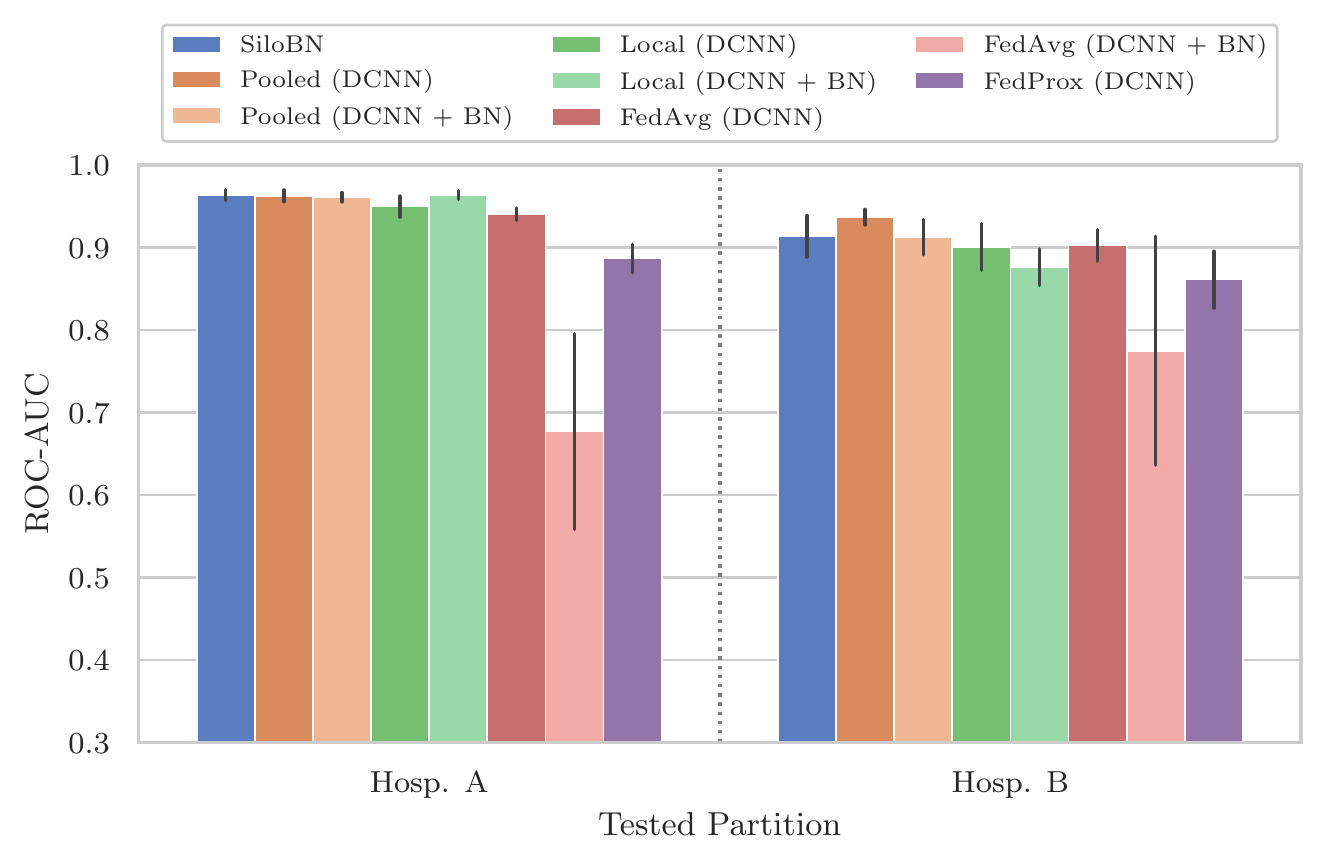}
        \includegraphics[width=0.7\textwidth, trim={0.2cm, 0.3cm, 0.2cm, 0.2cm}, clip]{%
        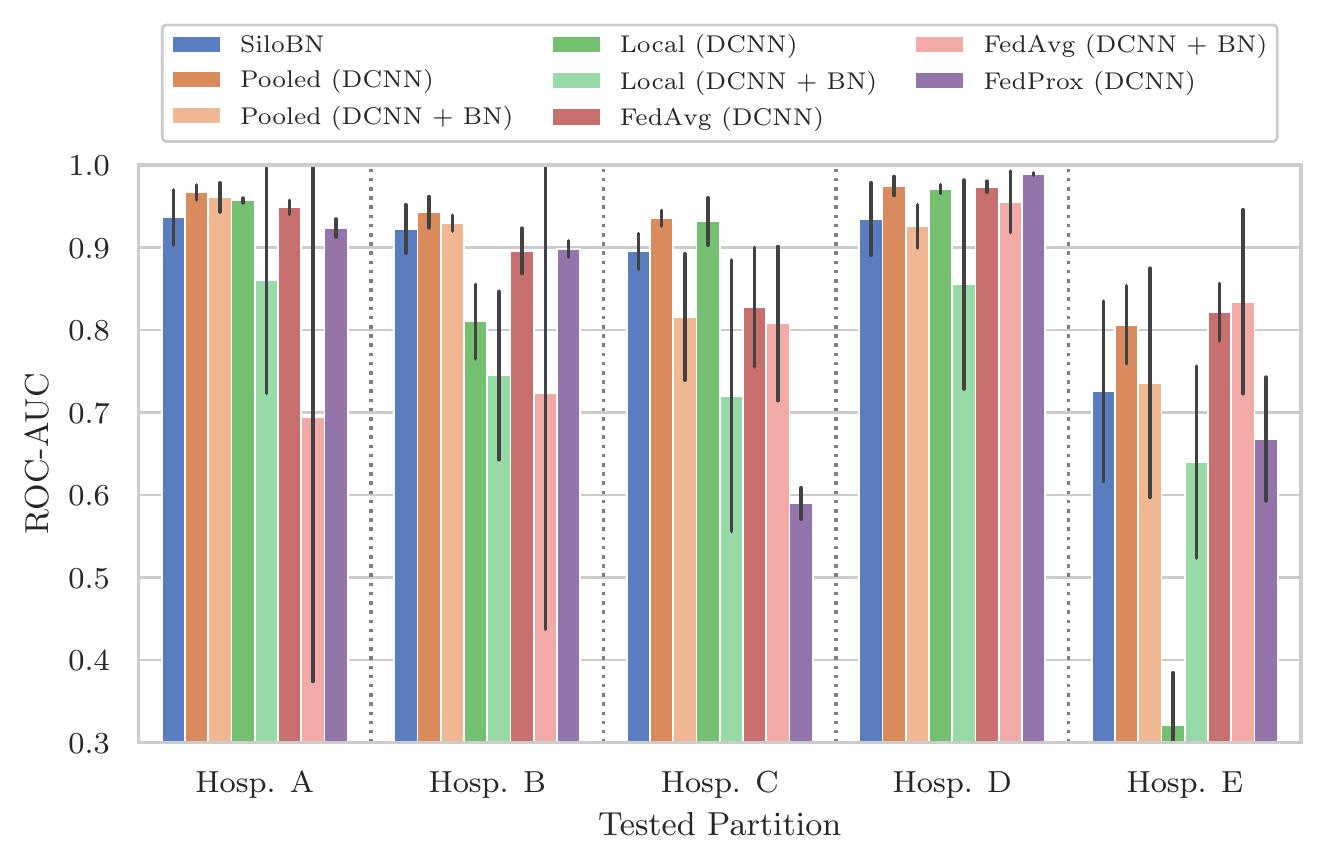}
\caption{\label{fig:c16c17siloedbndetailed}%
    Predictive performance of different training approaches used for training DCNN with and without
    BN layers. Performance is measured on per-center held-out test sets.
    Note that the \textit{Pooled}, \textit{FedProx}, and \textit{FedAvg} approaches
    produce a single model which is evaluated on multiple test sets, while the \textit{Local} and \textit{SiloBN}
    approaches produce one model per center, which is then evaluated on its corresponding center-specific test set.
}
\end{center}
\end{figure}

\paragraph{Training setting}
For tile classification, we use two deep convolutional neural network (DCNN)
architectures that only differ by the presence or absence 
of BN layers, which we denote as DCNN and DCNN+BN, respectively.
These architectures are detailed
in Fig.~\appendixcontext{\ref{fig:architecture}}{3} in Supp. Mat.
Note that, due to computational constraints,
this architecture is shallower than standard
ones~\cite{he2016residual,hu2018squeeze,tan2019efficient},
yielding better performance without BN layers,
which is not entirely realistic.
Each FL algorithm is run with $E=1$ or $E=10$ local batch
updates.
Each training session is repeated 5 times using different 
random seeds for weight initialization and data ordering.
Finally, all local optimization is 
performed using Adam~\cite{KB14}.
Tab.~\appendixcontext{\ref{table:hyperparameters}}{4} in Supp. Mat. provides
all related training hyperparameters.

\paragraph{Evaluation}
To compare the different methods, we measure intra-center generalization performance 
by computing the area under the curve (AUC) of the 
ROC curve of the trained model's predictions on each center's held-out 
data for single-model
methods (\textit{Pooled}, \textit{FedAvg}, or \textit{FedProx}), and 
personalized models are tested on held-out data for their specific training domain 
(\textit{Local} and \textit{SiloBN}).
We also report mAUC, which 
corresponds to the pan-center mean of intra-center generalization AUCs.
\begin{table}
    \centering
    \caption{Mean AUC over federated centers for
    different training approaches on FL-C16 and FL-C17.
    The proposed \textit{SiloBN} is on par or better
    than other FL methods, and is shown to be
    the most effective approach when using BN layers.
    \label{tab:meanAUC}
    }
    {
    \renewcommand{\arraystretch}{1.1}
    \setlength{\tabcolsep}{3pt}
    \begin{tabular}{lccc}
        \multicolumn{4}{c}{\textsc{With BN Layers}}\\
        \toprule
        {} & {$E$} & {FL-C16} & {FL-C17} \\
        \cline{2-4}
        {\emph{Pooled}}           & {--} & $0.94\pm 0.03$ & $0.87 \pm 0.11$  \\
        {\emph{Local}}          & {--} & $0.92 \pm 0.05 $ & $0.76 \pm 0.16 $ \\
        \cline{2-4}
        {\bfseries \emph{SiloBN}} & {1} & $0.94 \pm 0.03$ &  $0.86 \pm 0.13$\\
                                       & {\bfseries 10} & $\mathbf{0.94 \pm 0.03}$ & $\mathbf{0.88 \pm 0.10}$\\
        {\emph{FedAvg}} & {1} &  $0.81 \pm 0.05$ &  $0.70 \pm 0.18$\\
                                  & {10} & $ 0.73 \pm 0.14$ & $0.80 \pm 0.22$\\
        \bottomrule
    \end{tabular}
    }
\!
    {
    \renewcommand{\arraystretch}{1.1}
    \setlength{\tabcolsep}{3pt}
    \begin{tabular}{lccc}
        \multicolumn{4}{c}{\textsc{Without BN Layers}}\\
        \toprule
        {} & {$E$} & {FL-C16} & {FL-C17} \\
        \cline{2-4}
        {\emph{Pooled}}           & {--} & $0.95\pm 0.02$ & $0.93 \pm 0.07$  \\
        {\emph{Local}}          & {--} & $0.93 \pm 0.03 $ & $0.80 \pm 0.25 $ \\
        \cline{2-4}
                                        & & & \\
        {\emph{FedProx}} & {10} & $0.87 \pm 0.03 $ &  $0.81 \pm 0.16$\\
        {\bfseries \emph{FedAvg}} & {1} & $0.62 \pm 0.15$ & $0.80 \pm 0.17$ \\
                                                 & {\bfseries 10} & $\mathbf{0.92 \pm 0.02} $ & $\mathbf{0.89 \pm 0.07}$ \\
        \bottomrule
    \end{tabular}
    }
\end{table}

\paragraph{Impact of Federated Approaches}
Fig.~\ref{fig:c16c17siloedbndetailed}
provides per-center testing performance for $E=10$ updates on FL-C16 and FL-C17
and Tab.~\ref{tab:meanAUC} shows a full table of results averaged across
centers,
including a comparison of performance
over the number of local steps per federated round.
On both datasets, the proposed \textit{SiloBN} outperforms other FL strategies
for the DCNN+BN network,
and is on par with the pooled and local trainings.
For the DCNN network, for which \textit{FedAvg} and \textit{SiloBN} are equivalent,
we can make the same observation.
However, the lack of BN layers increases the sensitivity of \textit{FedAvg} to $E$.
We note that for FL-C17, some reported results in Fig.~\ref{fig:c16c17siloedbndetailed}
have very large error bars or very low AUC.
More analysis is required to understand the source of these instabilities.
Since some of these odd results occur for local trainings, they could e.g.
stem from poor local data quality.
Overall, from this set of experiments, we conclude that the proposed \textit{SiloBN}
is the only method that can fully utilize BN layers in the FL setting.
\subsection{Out-of-domain generalization experiments}
\label{sec:domaingeneralizationxps}
We now investigate the ability of the proposed method to transfer
to new domains.
We first train a model on FL-C16 and test it on FL-C17.
\begin{table}[t!]
\centering
\caption{Domain generalization of \textit{SiloBN} and \textit{FedAvg} (DCNN)
from FL-C16 to FL-C17. On average, \textit{SiloBN}
outperforms \textit{FedAvg} and
yields more stable results.
\label{tab:domaingen}
}
\sisetup{
    group-separator={,},
    group-minimum-digits=3,
    table-number-alignment = right,
    table-figures-integer = 6,
    detect-weight = true,
    detect-inline-weight = math
}
\resizebox{\columnwidth}{!}{%
\begin{tabular}{lccccc|c}
\toprule
    Hospital & A & B & {C} & {D} & {E} & {\emph{Mean}} \\
\cmidrule{2-7}
    \textit{SiloBN}+\textit{AdaBN} & $\;0.94 \pm 0.01$ & $\mathbf{\;0.91 \pm 0.04}$  & $\;0.93 \pm 0.01$  & $\mathbf{\;0.98 \pm 0.01}$ & $\mathbf{\;0.95 \pm 0.01}$ & $\mathbf{\;0.94 \pm 0.02}$ \\
    \textit{FedAvg} (DCNN) & $\mathbf{\;0.95 \pm 0.02}$ & $\;0.86 \pm 0.03$  & $\mathbf{\;0.95 \pm 0.01}$  & $\mathbf{\;0.98 \pm 0.02}$ & $\;0.88 \pm 0.03$ & $\;0.92 \pm 0.05$ \\
\bottomrule
\end{tabular}
}
\end{table}
We compare the proposed \textit{SiloBN}, where testing is done with \textit{AdaBN} as
explained in Sec.~\ref{sec:siloedbn},
to a DCNN without BN layers trained with \textit{FedAvg},
for which no adaptation is needed.
Per-hospital generalization results are provided
in Tab.~\ref{tab:domaingen}.
We note that the proposed algorithm outperforms the \textit{FedAvg}-trained DCNN model
on average across centers.
Moreover, we note that its generalization performance is much
more stable, both across
trainings (lower variance) and across centers.
\section{Conclusion}\label{sec:conclusion}
We have introduced \textit{SiloBN}, a novel FL strategy which relies on
\textit{local-statistic} BN layers in DCNNs.
Experiments on real-world multicentric histopathology
datasets have shown that this method yields similar, or better,
intra-center generalization capabilities than existing FL methods.
Importantly, the proposed approach outperforms existing methods on
out-of-domain generalization in terms of performance and stability,
while enjoying a better privacy profile 
by not sharing local activation statistics.
Future works could quantify the privacy benefits of this approach
as well as study its applicability to other domains, e.g. radiology.
\bibliographystyle{splncs04}
\bibliography{references}
\appendix
\setcounter{figure}{2}
\setcounter{table}{2}
\clearpage
\section{Supplementary Material}
\begin{table}[h!]
\centering
\sisetup{
    group-separator={,},
    group-minimum-digits=3,
    table-number-alignment = right,
    table-figures-integer = 6,
    detect-weight = true,
    detect-inline-weight = math
}
\resizebox{\columnwidth}{!}{%
    \begin{tabular}{l *{3}{S} c@{\hskip 5pt} *{6}{S[table-figures-integer=5]}}
    \toprule
\multicolumn{10}{l}{\textbf{Training sets}} \\
    & \multicolumn{3}{c}{\bfseries FL-C16} & &  \multicolumn{6}{c}{\bfseries FL-C17} \\
    Hospital & A & B & {\emph{Sum}} & & A & B & {C} & {D} & {E} & {\emph{Sum}} \\
\cmidrule{2-4} \cmidrule{6-11}
    Healthy & 50892 & 40000 & 90892 &  & 4562 & 3500 & 4900 & 3900 & 3700 & 20562 \\
    Tumor & 27078 & 22981 & 50059 & &  1740 & 597 & 3046 & 2895 & 8856 & 17134\\
\cmidrule{2-4} \cmidrule{6-11}
    \emph{Sum} & 77970 & 62981 & \bfseries 140,951 & & 6302 & 4097 & 7946 & 6795 & 12556 & \bfseries 37,696 \\
\bottomrule
\multicolumn{10}{l}{\textbf{Validation sets}} \\
    & \multicolumn{3}{c}{\bfseries FL-C16} & &  \multicolumn{6}{c}{\bfseries FL-C17} \\
    Hospital & A & B & {\emph{Sum}} & & A & B & {C} & {D} & {E} & {\emph{Sum}} \\
\cmidrule{2-4} \cmidrule{6-11}
    Healthy & 9000 & 21565 & 30565 &  & 1000 & 1200 & 1500 & 1100 & 1500 & 6300 \\
    Tumor & 6292 & 7284 & 13576 & & 205 & 125 & 259 & 38 & 98 & 725 \\
\cmidrule{2-4} \cmidrule{6-11}
    \emph{Sum} & 15292  & 28849 & \bfseries 44,141 & & 1205 & 1325 & 1759 & 1138 & 1598 & \bfseries 7,025 \\
\bottomrule
\multicolumn{10}{l}{\textbf{Test sets}} \\
    & \multicolumn{3}{c}{\bfseries FL-C16} & &  \multicolumn{6}{c}{\bfseries FL-C17} \\
    Hospital & A & B & {\emph{Sum}} & & A & B & {C} & {D} & {E} & {\emph{Sum}} \\
\cmidrule{2-4} \cmidrule{6-11}
    Healthy & 20362 & 10000 & 30362 &  & 800 & 1100 & 1100 & 1000 & 900 & 4900 \\
    Tumor & 6115 & 9159 & 15274 & & 523 & 847 & 647 & 1088 & 912 & 4017 \\
\cmidrule{2-4} \cmidrule{6-11}
    \emph{Sum} & 26477 & 19159 & \bfseries 45,636 & & 1323 & 1947 & 1747 & 2088 & 1812 & \bfseries 8,917 \\
\bottomrule
\end{tabular}
}
\caption{\label{table:training_set_distribution}%
    Distribution of healthy and tumor-containing tile images
    across centers for each of the constructed federated datasets.
}
\end{table}
\begin{table}[h!]
\centering
\resizebox{0.65\columnwidth}{!}{
\begin{tabular}{|c|c|c|c|c|c|c|}
\toprule
Network & Method & $\beta_1$ & $\beta_2$ &
\begin{tabular}{c}
    $L2$ \\
    reg.
\end{tabular}
&
\begin{tabular}{c}
    FedProx \\
    reg.
\end{tabular} &
\begin{tabular}{c}
Learning\\rate
\end{tabular}
\\
\hline
DCNN + BN & 
All except
FedAvg
& 0 & 0.99 & 0.001 & -- & $10^{-4}$ \\
DCNN + BN & FedAvg & 0 & 0.999 & 0.1 & -- & $10^{-3}$ \\
\hline
DCNN & All except Fedprox & 0 & 0.99 & 0 & -- & $10^{-3}$ \\
DCNN & FedProx & 0 & 0.99 & 0 & 0.1 & $10^{-3}$ \\
\bottomrule
\end{tabular}
}
\resizebox{0.65\columnwidth}{!}{
\begin{tabular}{|c|c|c|c|c|c|}
\toprule    
Dataset & Network & Method & E & Batch size & Number of rounds \\
\hline
FL-C16 & All unless specified & -- & 1 & 32 & 21875 \\
FL-C16 & All unless specified & -- & 10 & 32 & 2187 \\
FL-C16 & DCNN+BN & Pooled & -- & 64 & 21875 \\
FL-C16 & DCNN & FedAvg  & 1 & 32& 5000 \\
FL-C16 & DCNN & FedProx & 10 & 32 & 300 \\
\hline
FL-C17 & All unless specified & -- & 1 & 32 &2355 \\
FL-C17 & All unless specified & -- & 10 & 32 & 235 \\
FL-C17 & DCNN+BN & Pooled & -- & 160 & 2355 \\
FL-C17 & DCNN & FedAvg & 1 & 32&500 \\
FL-C17 & DCNN & FedProx & 10 & 32 &50 \\
\bottomrule
\end{tabular}
}  %
\caption{Hyperparameters used in our experiments obtained with the
validation set.
When applicable, the \texttt{momentum} of the BN layers is set to $0.1$
and the value \texttt{eps} to $10^{-5}$.
The first and second moment decay hyperparameters of 
the Adam optimizer are noted $\beta_1$ and $\beta_2$.
For each dataset, local training used parameters corresponding to $E=1$.
}
\label{table:hyperparameters}
\end{table}
\begin{figure}[ht!]
    \centering
    \includegraphics[width=0.6\linewidth]{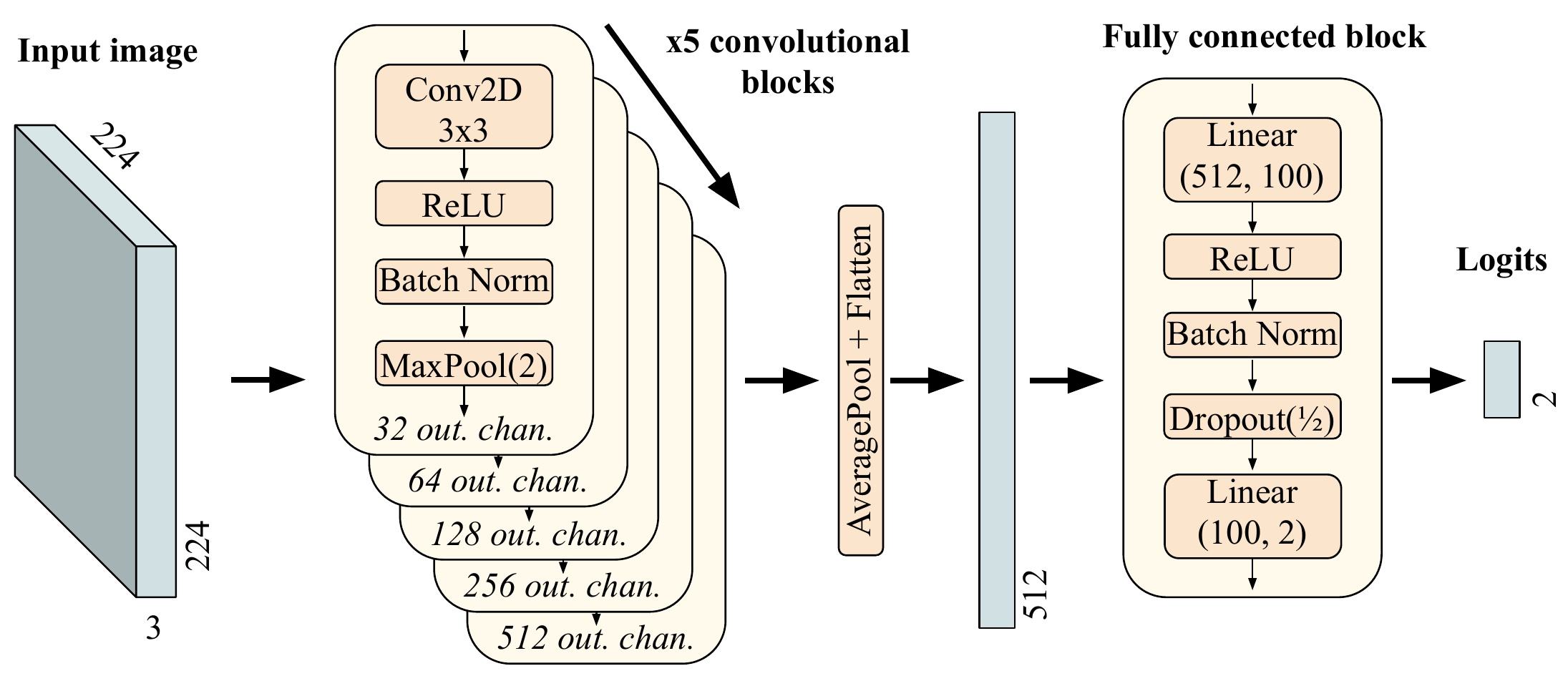}
    \caption{Architecture of the deep convolutional neural network
    used in all the experiments. All convolutional layers use
    zero-padding of size 1 on spatial dimensions.
    The number of output channels of each
    convolutional layer corresponds to the number of output channels
    (abbreviated as \textit{out. chan.}) of the corresponding
    convolutional block.
    All convolutional and linear layers include a bias.
    \label{fig:architecture}
    }
\end{figure}
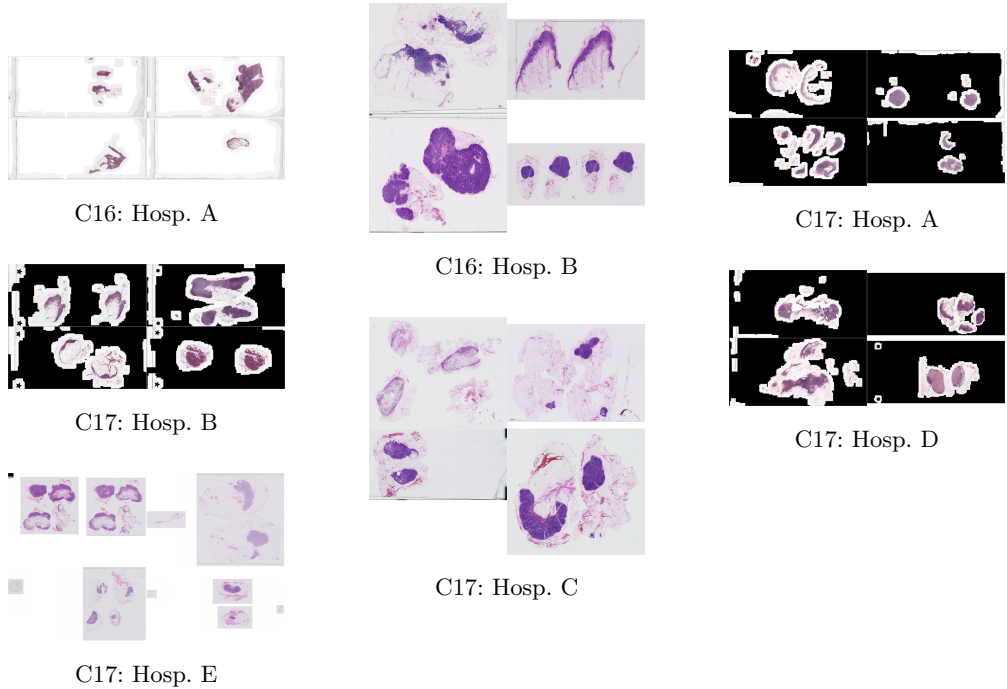
\begin{figure}
    \begin{tikzpicture}
        \node (C16HA) [] {\includegraphics[width=0.3\textwidth]{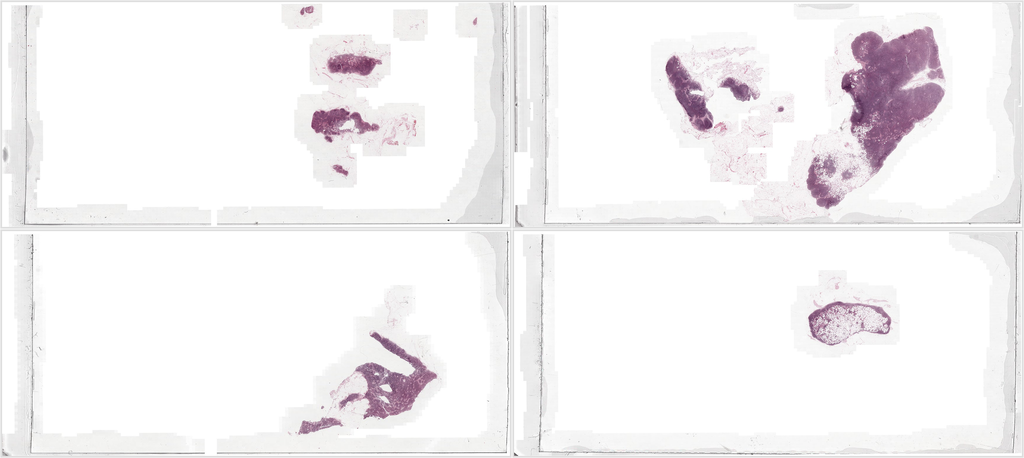}};
        \node (cC16HA) [below = 0.1cm of C16HA] {C16: Hosp. A};

        \node (C16HB) [right = 0.9cm of C16HA] {\includegraphics[width=0.3\textwidth]{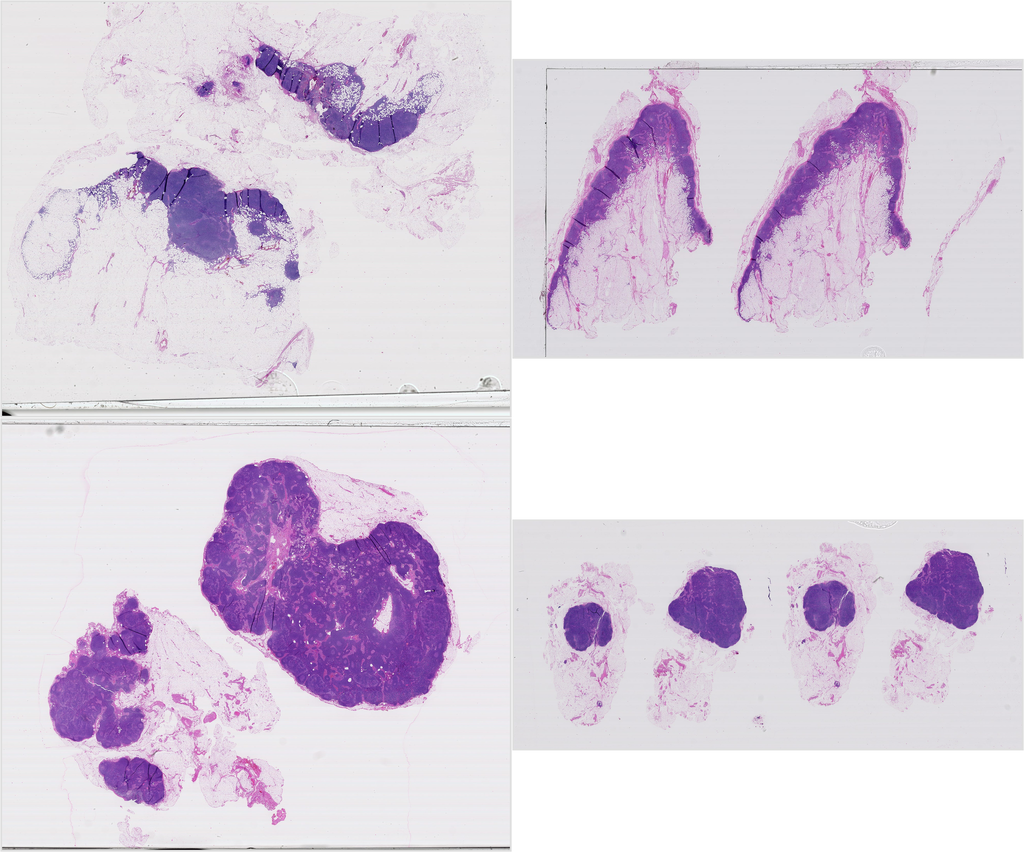}};
        \node (cC16HB) [below = 0.1cm of C16HB] {C16: Hosp. B};

        \node (C17HA) [right = 0.9cm of C16HB] {\includegraphics[width=0.3\textwidth]{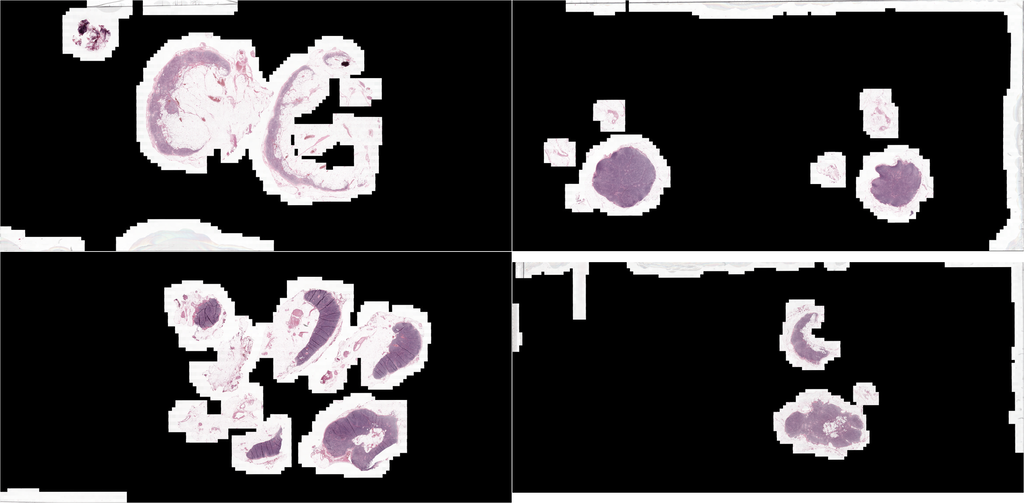}};
        \node (cC17HA) [below = 0.1cm of C17HA] {C17: Hosp. A};

        \node (C17HB) [below = 0.9cm of C16HA] {\includegraphics[width=0.3\textwidth]{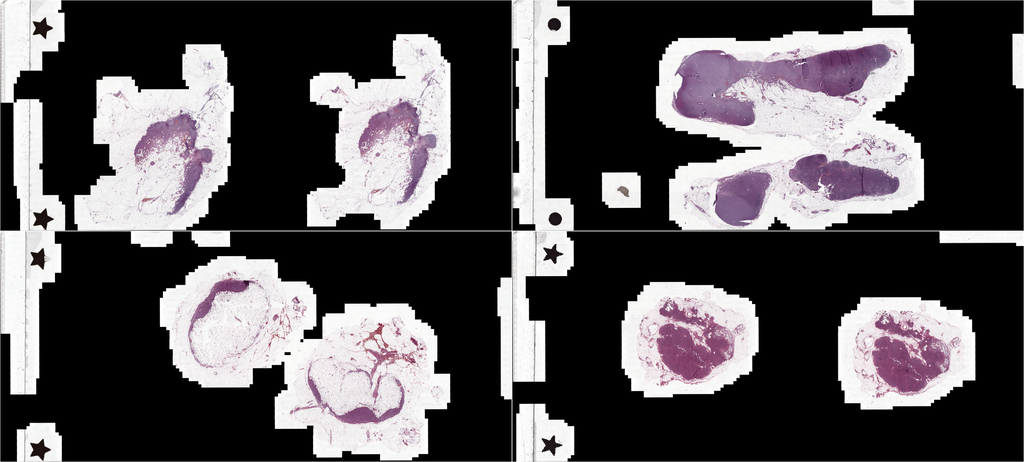}};
        \node (cC17HB) [below = 0.1cm of C17HB] {C17: Hosp. B};

        \node (C17HC) [below = 0.9cm of C16HB] {\includegraphics[width=0.3\textwidth]{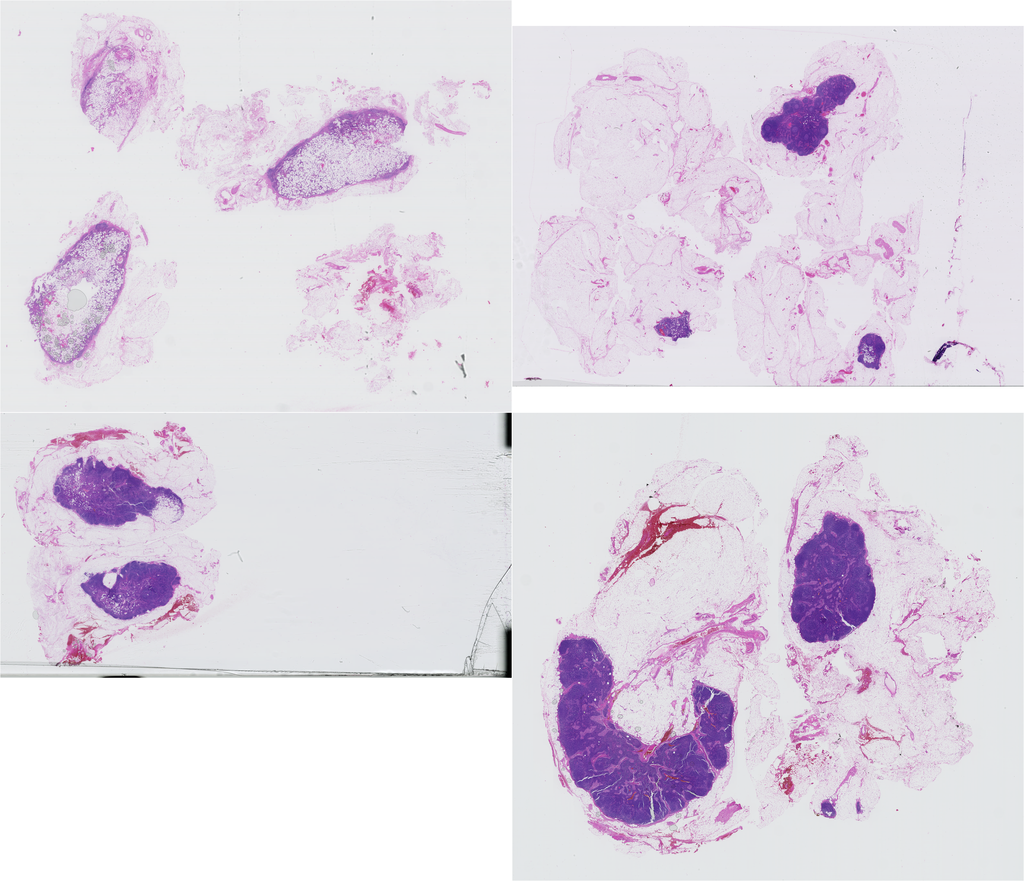}};
        \node (cC17HC) [below = 0.1cm of C17HC] {C17: Hosp. C};
    
        \node (C17HD) [below = 0.9cm of C17HA] {\includegraphics[width=0.3\textwidth]{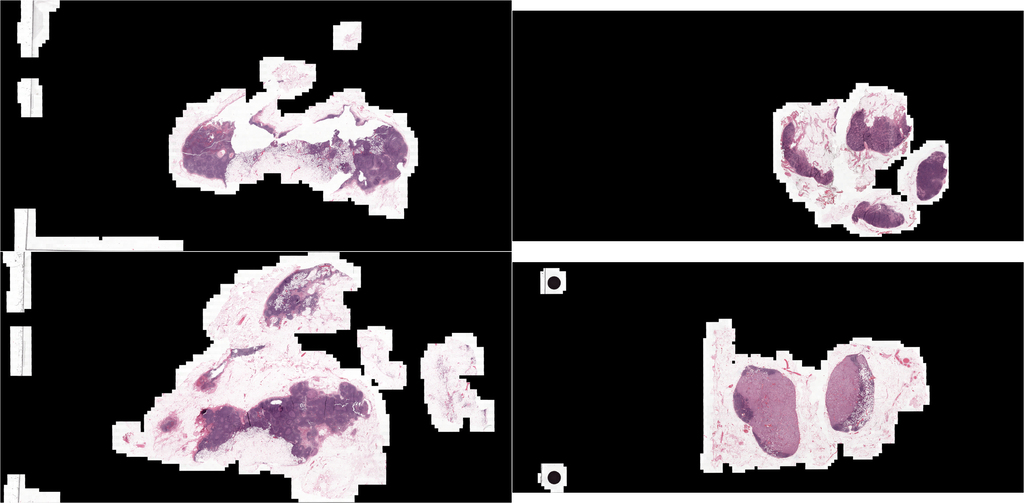}};
        \node (cC17HD) [below = 0.1cm of C17HD] {C17: Hosp. D};

        \node (C17HE) [below = 0.9cm of C17HB] {\includegraphics[width=0.3\textwidth]{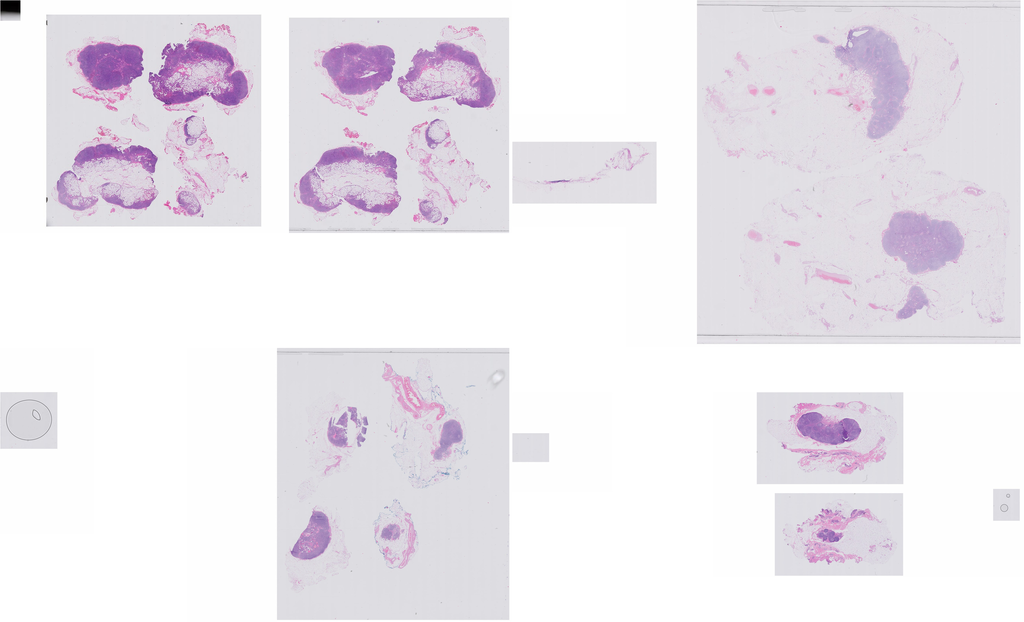}};
        \node (cC17HE) [below = 0.1cm of C17HE] {C17: Hosp. E};
    \end{tikzpicture}
    \caption{\label{fig:cam-slides}%
        Four representative WSIs for each identified center in the Camelyon 16 and Camelyon 17 datasets.
    }
\end{figure} 
\end{document}